\documentclass{article}
\usepackage{ijcai26}

\usepackage{times}
\usepackage{soul}
\usepackage{url}
\usepackage[hidelinks]{hyperref}
\usepackage[utf8]{inputenc}
\usepackage[small]{caption}
\usepackage{graphicx}
\usepackage{amsmath}
\usepackage{amsthm}
\usepackage{booktabs}
\usepackage{algorithm}
\usepackage[switch]{lineno}

\usepackage{booktabs}
\usepackage{multirow}
\usepackage[normalem]{ulem}
\useunder{\uline}{\ul}{}
\usepackage{graphicx}
\usepackage{subcaption}
\usepackage{makecell}
\usepackage[table,xcdraw]{xcolor}

\usepackage{amsmath} 
\usepackage{amsfonts} 
\usepackage{algorithm} 
\usepackage{algpseudocode} 

\title{NEST: Tackling Dataset-Level Distribution Shifts via Regime-Oriented Mixture-of-Experts}

\author{
Lanhao Li$^1$
\and
Bingshu Xie$^1$\and
Lijun Sun$^1$\and
Xin Xue$^1$\and
Haoyi Zhou$^{1}$\footnote{Corresponding Author}\And
Jianxin Li$^1$
\affiliations
$^1$Beihang University\\
\emails
\{lilanhao, xiebingshu, sunlijun1, xuexin0102, haoyi, lijx\}@buaa.edu.cn
}

\begin{document}

\maketitle

\begin{abstract}
     Accurate long-term forecasting in complex systems is frequently compromised by dataset-level distribution shifts, where diverse underlying behavioral modes and evolving system states drive the dynamic multivariate time-series. While existing methods predominantly focus on local temporal shifts, they fail to explicitly model the global structural challenge where datasets are composites of distinct operational regimes. In this paper, we propose NEST, a specialized framework designed to model and recompose these evolving structures through a two-phase dense MoE architecture. NEST first facilitates structural specialization by partitioning the dataset into distinct operational regimes through unsupervised clustering in a principled moment-entropy space. We introduce a regime-oriented router mechanism that generates initial expert weights based on temporal content, subsequently refined through geometric modulation to regime centroids. Crucially, rather than acting as monolithic predictors, individual experts function as specialized kernels that capture regime-specific dynamics by evolving unique variate-attention patterns. Extensive evaluations on diverse benchmarks, including heterogeneous network traffic and physical phenomena, demonstrate that NEST consistently achieves state-of-the-art performance. Our code and datasets are available at \url{https://github.com/Aaralshin/NEST}.
\end{abstract}

\section{Introduction}
\label{sec:introduction}
\begin{figure}[t]
    \centering  
    \includegraphics[width=0.45\textwidth]{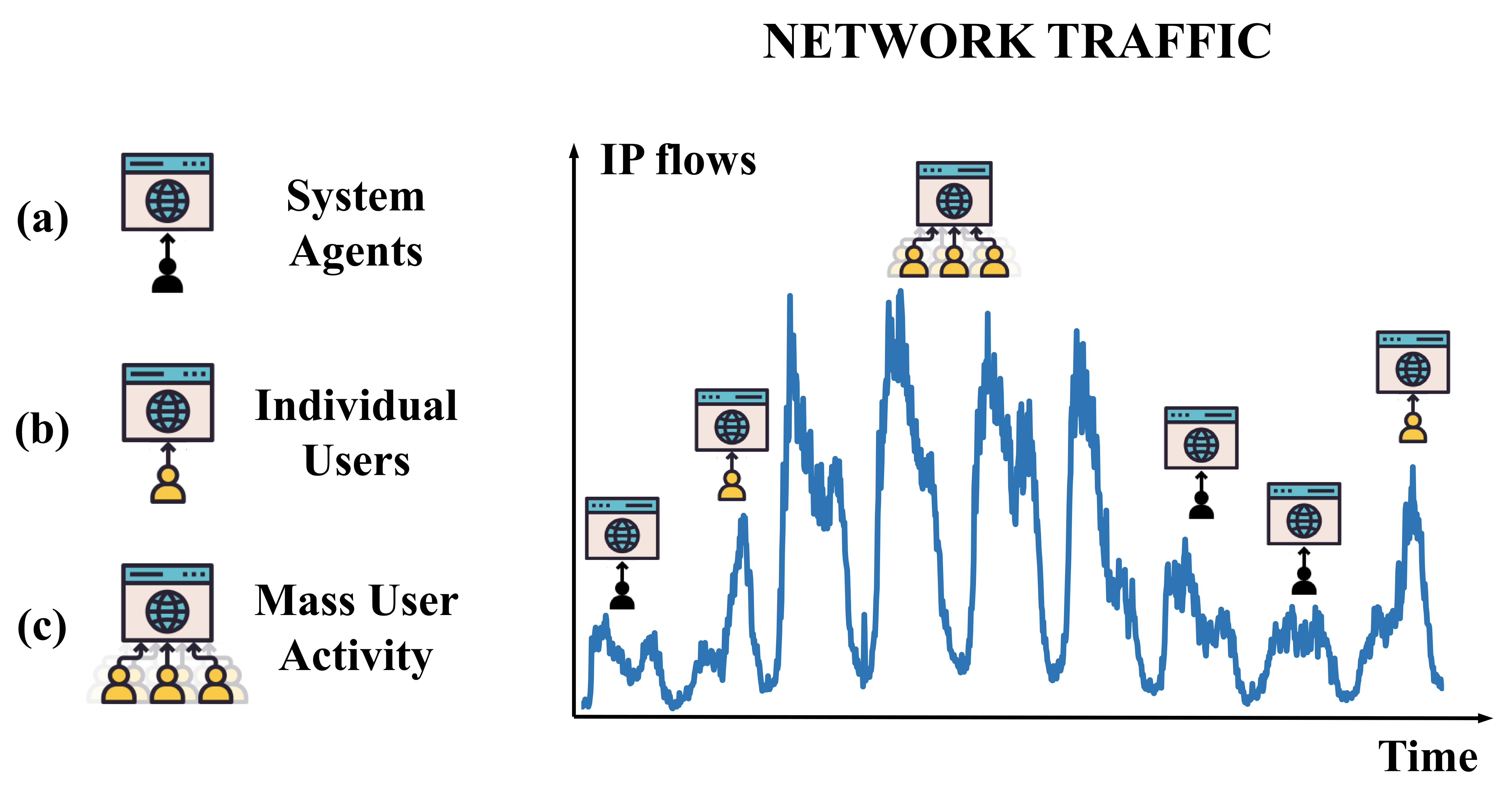}
    \caption{Conceptual illustration of dataset-level distribution shift. Using network traffic as an example, the IP flow volume fluctuates over time according to different underlying behavioral modes.}
    \label{fig:Conceptual illustration}
\end{figure}

Time series forecasting serves as a cornerstone for understanding and managing complex systems, ranging from the global web infrastructure~\cite{basu1996time} and power grids~\cite{boehme2007applying} to intricate physical phenomena like space weather dynamics~\cite{pulkkinen2007space}. In these domains, the ability to provide accurate long-term predictions is essential for proactive resource allocation, anomaly detection, and ensuring system stability. However, real-world data streams are challenging to model due to their high non-stationarity and significant distribution shift across datasets.

This distribution shift is not random; it is a manifestation of diverse underlying behavioral modes and evolving system states and can be recurrent during a long period. For example, as illustrated in Figure~\ref{fig:Conceptual illustration}, this heterogeneity is driven by different underlying behavioral modes, each generating a traffic pattern with a distinct statistical signature. Icon (a) illustrates low-volume, baseline traffic might be generated by automated system check-ins or background processes; icon (b) illustrates standard daily peaks, which could correspond to individual users engaging in typical interactive sessions; icon (c) illustrates the highest traffic surges, often reflecting coordinated, large-scale activity, such as many users accessing a popular service simultaneously during a major event. A monolithic forecasting model, when trained on data containing this mixture of behaviors, is forced to learn a single, ``averaged" representation that fails to accurately capture the specific dynamics of any individual mode, leading to suboptimal performance. 

Despite its prevalence, few existing methods address this dataset-level shift. The predominant focus remains on mitigating the local temporal shift between the historical look-back window and the future prediction window. While valuable, this paradigm primarily corrects for evolving non-stationarity within a small temporal vicinity; it does not explicitly model the global, structural challenge where the dataset itself is a composite of distinct operational regimes.

To address this fundamental problem, we argue for a systematic approach rooted in the identification of distinct operational regimes. We hypothesize that the diverse dynamic behaviors within a dataset can be effectively differentiated by moment-entropy metrics. Specifically, we employ low-order statistical moments (mean and variance) to capture distribution intensity, alongside Singular Value Decomposition Entropy (SVDEn) to quantify structural complexity. By mapping data slices into this moment-entropy space, we apply unsupervised clustering to automatically partition the dataset into functionally distinct categories. This categorization ensures that specialized experts can be trained to master the specific patterns inherent to each regime, effectively capturing the dynamic reorganization of inter-variable dependencies that occurs as the system transitions between different types of behavior.

To this end, we propose NEST: Tackling Dataset-Level Distribution Shifts via Regime-Oriented Mixture-of-Experts. NEST is a two-phase MoE framework designed to address dataset-level distribution shifts (DDS) by explicitly modeling the transitions between underlying system regimes. The process begins with unsupervised regime discovery, where we identify distinct operational modes through clustering in the moment-entropy space. Based on these discovered regimes, we then execute a structured two-phase training protocol to cultivate a heterogeneous expert pool. In \textit{Phase 1}, we facilitate structural specialization within a heterogeneous expert pool; here, individual experts learn to model regime-specific sequence dynamics by evolving unique variate-attention patterns that capture the coupling logic of each mode. In \textit{Phase 2}, the router generates an initial weight based on temporal content, which is then refined through static topological modulation to anchor the decision via geometric proximity to regime centroids. By dynamically recomposing these specialized kernels, NEST effectively transforms the challenge of distribution shifts into a manageable process of regime identification and dependency reconfiguration.

The key contributions of our work are as follows:

\begin{itemize}

\item \textbf{Structural Regime Modeling:} We propose \textbf{NEST}, a two-phase MoE framework that models the evolution of complex system regimes by dynamically recomposing inter-variable dependencies. By decoupling evolving dynamics into specialized kernels, NEST effectively captures the structural transitions between operational modes that drive dataset-level distribution shifts.

\item \textbf{Principled Discovery and Routing:} We introduce an unsupervised discovery module using moment-entropy metrics to provide a mathematical basis for regime partitioning. Complementarily, we design a regime-oriented router that achieves precise expert orchestration by fusing expert weights initialization with distance-aware geometric modulation.

\item \textbf{Empirical Validation and Interpretability:} NEST achieves state-of-the-art (SOTA) performance across diverse benchmarks, including heterogeneous network traffic and multi-year physical phenomena. Extensive analysis of attention polymorphism and weight permutation further validates its unique capability in the structural recomposition of variable dependencies.

\end{itemize}

\section{Related Work}

\subsection{Deep Learning for Time Series Forecasting} \label{ssec:forecasting_related_work}

Early efforts leveraged RNNs like LSTM~\cite{siami2018comparison} and GRU~\cite{weerakody2021review} for temporal gating, though sequential constraints hindered long-horizon efficiency. The transition to Transformers~\cite{wu2020deep} introduced global dependencies, yet quadratic complexity prompted sparse or hierarchical variants like Informer~\cite{zhou2021informer} and Pyraformer~\cite{liu2021pyraformer}. Subsequent innovations focused on domain-specific kernels: Autoformer~\cite{wu2021autoformer} utilized auto-correlation, while FEDformer~\cite{zhou2022fedformer} applied frequency-domain attention via Fourier/wavelet transforms~\cite{torrence1998practical,cleveland1990stl}. Recent paradigms emphasize structural rethinking: PatchTST~\cite{nie2022time} processes sub-series patches for local semantics, while iTransformer~\cite{liu2023itransformer} inverts the architecture to model multivariate correlations by treating whole variates as tokens. Concurrently, DLinear~\cite{zeng2023transformers} demonstrated the efficacy of simple MLP-based decomposition. More recently, unified backbones like UniTS~\cite{gao2024units} have emerged as foundation models, achieving robust zero-shot generalization across diverse forecasting tasks.

\subsection{Handling Distribution Shifts}
\label{ssec:Handling Distribution Shifts}
Recent studies address non-stationarity in time series by adapting normalization or learning strategies at the \emph{instance} or \emph{window} level. Representative methods include instance-wise~\cite{ogasawara2010adaptive} learnable input normalization~\cite{passalis2019deep}, and reversible instance normalization (RevIN~\cite{kim2021reversible}), which removes non-stationary components during training and restores them during inference. Dish-TS\cite{fan2023dish} proposes a general neural paradigm that mitigates distribution shift by employing a dual framework to separately model the distributions of the input and output spaces. FAN\cite{ye2024frequency} performs normalization by leveraging frequency-domain information. Subsequent work extends these ideas through slice-based normalization, such as SAN~\cite{liu2023adaptive}, where each slice is defined within the look-back window and the forecast horizon. However, as these methods focus on modeling distribution shifts between the look-back and forecast windows, they exhibit strong adaptability to short-term fluctuations but limited generalization under prolonged or regime transitions within the dataset, as they overlook the discrepancies induced by variations across training instances.

\section{Preliminaries}
\label{sec:preliminaries}

\subsection{Problem Definition}
\label{ssec:problem_definition}

The primary task addressed in this paper is long-term, multivariate time series forecasting. Given a historical multivariate time series, the goal is to learn a forecasting function $f(\cdot)$ that accurately predicts a sequence of future values based on a window of past observations.

Formally, let the input be a look-back window of length $L$, denoted as $\mathbf{x} = (x_1, x_2, \dots, x_L)$, where each multivariate observation $x_t \in \mathbb{R}^d$ is a $d$-dimensional vector at time step $t$. The model must learn a function $f$ that maps the input window to the subsequent $H$ time steps, where $H$ is the prediction horizon:
\begin{equation}
    \hat{\mathbf{y}} = f(\mathbf{x})
\end{equation}
Here, $\hat{\mathbf{y}} = (\hat{y}_1, \hat{y}_2, \dots, \hat{y}_H)$, where each $\hat{y}_t \in \mathbb{R}^d$, is the predicted sequence. The objective is to minimize the discrepancy between the prediction $\hat{\mathbf{y}}$ and the ground truth future sequence $\mathbf{y} = (y_1, y_2, \dots, y_H)$.


The fundamental challenge we address is dataset-level distribution shift (DDS), where a long-term time series is composed of multiple, heterogeneous operational regimes. Unlike local non-stationarity, DDS reflects the structural evolution of the system over extended horizons. For instance, in ionospheric total electronic content(TEC) data, the decadal solar cycles introduce profound shifts in signal characteristics that span years; similarly, in network traffic, the transition from background maintenance to coordinated user surges represents a fundamental change in system behavior. Mathematically, a monolithic function $f$ struggles in this scenario because the mapping between history and future is not invariant. Instead, the underlying inter-variable dependencies are dynamically reorganized as the system transitions between regimes. When a single model is forced to minimize a global objective across this mixture of distinct behaviors, it inevitably learns an ``averaged" representation. This averaging effect obscures the specific coupling logic of individual regimes, leading to a loss of fidelity in capturing the precise dynamics required for robust long-term forecasting. To resolve this, it is necessary to identify these regimes and master their unique patterns via specialized kernels.

\begin{figure*}[ht!]
    \centering  
    \includegraphics[width=0.95\textwidth]{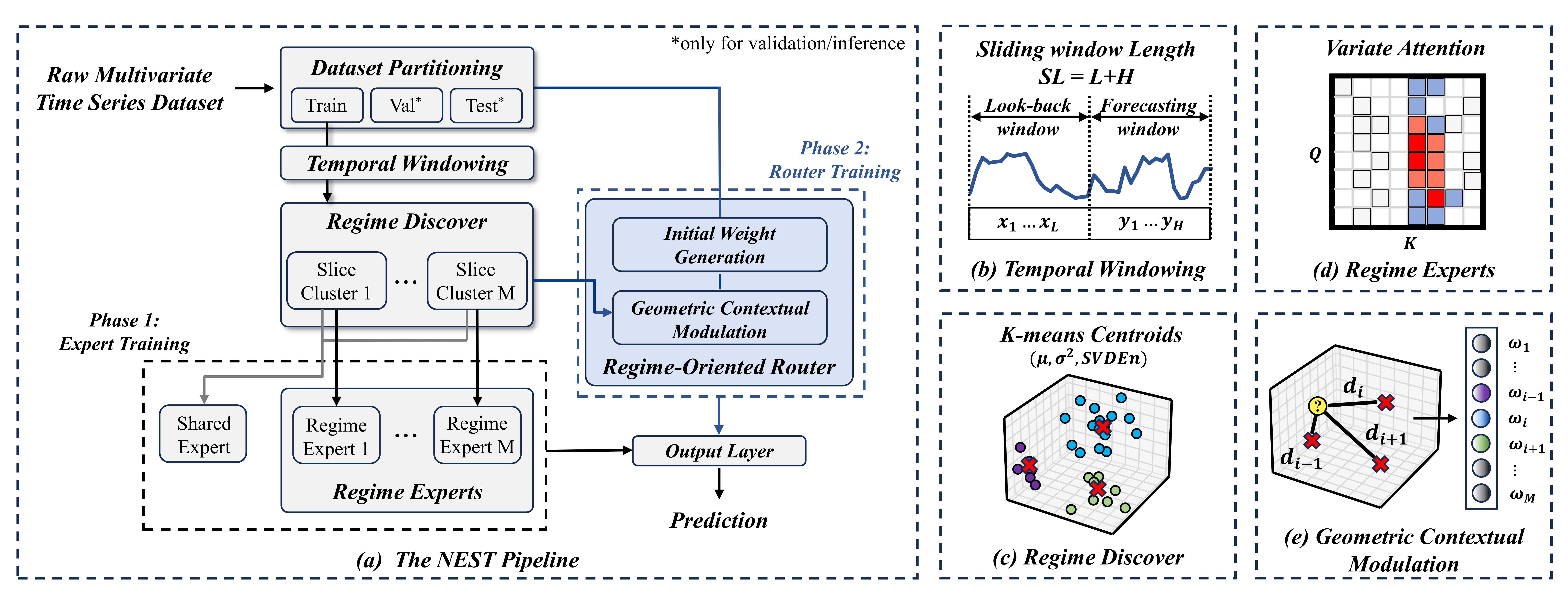}
    \caption{\textbf{Overview of the proposed NEST framework and its key components.}
        \textbf{(a) The NEST Pipeline:} Shows the complete two-phase pipeline. \textbf{Phase 1} performs regime discovery and expert training. \textbf{Phase 2} freezes the experts and trains the regime-oriented router to adaptively combine their outputs for the final prediction.
        \textbf{(b) Temporal Windowing:} Illustrates the process of creating data slices of length $SL$ from the raw time series.
        \textbf{(c) Regime Discovery:} A conceptual visualization of the regime discovery process, where data slices are clustered in a feature space defined by their statistical and complexity features.
        \textbf{(d) Regime Expert:} Each regime expert consists of a variate-attention mechanism to model inter-variate relations.
        \textbf{(e) Geometric Contextual Modulation:}  Modulates expert weights based on the feature-space distance of an input sample to the pre-computed regime centroids after the initialization of the weights.
    }
    \label{fig:arch}
\end{figure*}

\section{Methodology}
\subsection{Overall Framework}
\label{ssec:framework}

To address the challenge of dataset-level distribution shift, we propose NEST: Tackling Dataset-Level Distribution Shifts via Regime-Oriented Mixture-of-Experts. The framework is built upon the principle of regime-specific structural modeling, where the heterogeneous operational modes within a long-term time series are first identified and then masterfully captured by a specialized expert pool. Instead of learning a single averaged representation, NEST anchors its predictive logic in the distinct dynamical patterns inherent to different system states.

The NEST architecture comprises three key components: 
\begin{itemize} 
    \item \textbf{Unsupervised Regime Discovery:} A principled module that partitions the dataset into functionally distinct regimes. By employing \textbf{moment-entropy metrics}, it characterizes each data segment through its statistical distribution and structural complexity, ensuring a robust mathematical basis for regime identification.

    \item \textbf{Specialized Expert Pool:} A heterogeneous ensemble consisting of specialized ``regime experts" and a ``shared expert". Each regime expert is trained to achieve structural specialization, evolving unique variate-attention patterns to capture the specific inter-variable dependencies of its assigned regime, while the shared expert retains global, overarching knowledge.

    \item \textbf{Regime Oriented Router:} A dynamic gating network that orchestrates the expert pool through a functional division of labor. The router achieves precise expert selection by integrating content-based temporal regime alignment with geometric contextual modulation, enabling the model to accurately recompose specialized patterns in response to evolving system dynamics.
\end{itemize}

The NEST pipeline, illustrated in Figure~\ref{fig:arch}, follows a two-phase process. \textbf{Phase 1} consists of an unsupervised regime discovery step followed by the training of a specialized expert pool on the discovered data partitions. In \textbf{Phase 2}, these experts are frozen, and the router is trained to adaptively aggregate their predictions for context-aware forecasting.



\subsection{Unsupervised Regime Discovery via Data Clustering}
\label{ssec:clustering}

The foundational step of our NEST framework is the automated identification of distinct operational regimes within the time series dataset. This is achieved through an unsupervised clustering process performed as a pre-processing step before the main model training.

First, we transform the training time series into a set of overlapping instances, or ``slices". Each slice $\mathbf{s}_i$, starting at time index $i$, is defined as a sequence of length $SL = L+H$, where $L$ is the input window length and $H$ is the prediction horizon:
\begin{equation}
    \mathbf{s}_i = \{x_t\}_{t=i}^{i+SL-1}
\end{equation}

Next, for each slice $\mathbf{s}_i$, we extract a low-dimensional feature vector $\mathbf{v}_i$ designed to capture its core dynamic properties. This vector includes: \textbf{Moment Metric:} We compute the mean ($\mu$) and variance ($\sigma^2$) of the slice, representing its central tendency and volatility. \textbf{Entropy Metric:} To capture intrinsic complexity, we compute the Singular Value Decomposition Entropy (SVDEn). This involves constructing a trajectory matrix $\mathbf{X}$ from the slice $\mathbf{s}_i$ with an embedding dimension $d$:


    \begin{equation}
        \mathbf{X} = \begin{pmatrix} s_1 &  \dots & s_{SL-d+1} \\ \vdots &  \ddots & \vdots \\ s_d & \dots & s_{SL} \end{pmatrix} \in \mathbb{R}^{d \times (SL-d+1)}
    \end{equation}
We then perform SVD on $\mathbf{X}$ to obtain its singular values $\{X_{\sigma_{j}}\}_{j=1}^k$. These values are normalized to form a probability distribution $p_j = X_{\sigma_{j}}/ {\sum_{l=1}^k X_{\sigma_{l}} }$. The SVDEn is the Shannon entropy of this distribution:
    \begin{equation}
        \text{SVDEn}(\mathbf{s}_i) = - \sum_{j=1}^{k} p_j \log(p_j)
    \end{equation}

The final feature vector for each slice is the concatenation of these metrics: $\mathbf{v}_i = [\mu(\mathbf{s}_i), \sigma^2(\mathbf{s}_i), \text{SVDEn}(\mathbf{s}_i)]$. With the training set represented as a collection of feature vectors $\{\mathbf{v}_i\}$, we apply K-Means clustering to partition the data into $M$ clusters. The objective is to find the set of centroids $C = \{\mathbf{c}_1, \dots, \mathbf{c}_M\}$ that minimizes the within-cluster sum of squares. Each centroid $\mathbf{c}_m$ serves as a prototypical representation of its regime and is utilized by the router.

A crucial detail of our regime discovery process is the use of data slices spanning both the look-back window and the prediction horizon. This design choice ensures that the discovered regimes are defined by the complete dynamic behavior of a sequence, including its evolution into the future, thus creating more forecasting-oriented clusters. At inference time (i.e., during validation and testing), only the look-back window of length $L$ is available. To assign an incoming sequence to a regime, we compute its feature vector (mean, variance, SVDEn) using solely this available $L$-length data. We then operate under the assumption that the statistical properties of the look-back window serve as a sufficient proxy to identify the most likely regime for the complete $L+H$ dynamic. That is, from the historical look-back window to the forecasting window, the data distribution remains consistent, which can be ensured by the method discussed in Section \ref{ssec:Handling Distribution Shifts}. This procedure strictly avoids data leakage, as the ground truth future values are never used to compute features or determine regime assignment at inference time. The regime centroids are established exclusively from the training set, and the mapping during inference relies only on historical data.

\subsection{The Multi-Expert Pool}
\label{ssec:experts}
The predictive core of the NEST framework is a heterogeneous pool of expert models, each designed to capture different facets of the time series dynamics. This pool contains two types of experts: specialized regime expert sets and a single, generalist shared expert.

\textbf{Regime Experts:} Corresponding to the $M$ regimes discovered via moment-entropy metrics, the model contains $M$ specialized regime experts, denoted as $\{E_m\}_{m=1}^M$. Each expert $E_m$ is responsible for learning the specific dynamic properties of its assigned regime $m$. To effectively capture the inherent pattern of the time series, each expert is implemented as a variate-attention module specifically optimized to capture regime-specific inter-variable dependencies. Given an input sequence $\mathbf{x} \in \mathbb{R}^{L \times C}$, where $L$ is the look-back window and $C$ is the number of variables, the expert focuses on the spatial correlations across the channel dimension:
\begin{equation}
E_{m}(\mathbf{x}) = \text{Attention}_{\text{variate}}(\mathbf{x})
\end{equation}
This design ensures that each expert masters the dynamic reorganization of variables within its assigned regime. The resulting attention maps provide an explicit representation of how variable dependencies shift across different operational states.

\textbf{Shared Expert:} To capture global patterns common across all regimes, the pool includes a single shared expert, $E_s$, which remains active across different data slices throughout Phase 1 training and employs variate attention to provide a stable, foundational forecast.

\subsection{Regime-Oriented Router}
\label{ssec:router}
The core of NEST's adaptability is the regime-oriented router, which dynamically orchestrates the expert pool through a two-step process to ensure stable and precise expert selection under complex distribution shifts.

\subsubsection{Initial Weight Generation}
The router first dynamically generates a preliminary weight distribution based on the input's temporal content. For the $M$ regime experts, the input embedding $\mathbf{x}_{\text{emb}}$ is processed by a lightweight Transformer encoder and a regime-specific linear head, yielding the preliminary weight distribution:
\begin{equation}
    \mathbf{w}_{\text{init}} = \text{Softmax}(\text{Linear}_{\text{regime}}(\text{Encoder}(\mathbf{x}_{\text{emb}})))
\end{equation}
In parallel, the output of the same encoder is passed through a different linear head and a Sigmoid activation function to compute the weight of the shared expert, $w_s$:
\begin{equation}
    w_s = \text{Sigmoid}(\text{Linear}_{\text{shared}}(\text{Encoder}(\mathbf{x}_{\text{emb}})))
\end{equation}
Using a Sigmoid function allows the model to treat the shared expert as an independent, additive component, without competing directly with the specialized regime experts for the same weight budget.

\subsubsection{Geometric Contextual Modulation}
To anchor the content-driven decisions within the global data manifold, we modulate $\mathbf{w}_{\text{init}}$ using the input's proximity to regime centroids in the moment-entropy space. For an input slice $\mathbf{s}_i$, we calculate the Euclidean distance $d_m$ between its moment-entropy metric vector $\mathbf{v}_i$ and each regime centroid $\mathbf{c}_m$:
\begin{equation}
    d_m = ||\mathbf{v}_i - \mathbf{c}_m||_2
\end{equation}
The distance is converted into a similarity weight $\tilde{\omega}_m$ via an inverse quadratic function. A softening factor $\alpha \in (0, 1]$ is applied to facilitate smooth collaboration during regime transitions:
\begin{equation}
    \tilde{\omega}_m = \left( \frac{1}{1 + d_m^2} \right)^\alpha
\end{equation}
The final adaptive weights $w'_{m}$ are produced by the element-wise modulation of temporal relevance and geometric context:
\begin{equation}
    w'_{m} = w_{\text{init}, m} \cdot \tilde{\omega}_m
\end{equation}

\subsubsection{Final Prediction and Training.}
The modulated scores are renormalized to produce the final regime expert weights $\mathbf{w}_{\text{regime}}$. A separate gate computes the weight $w_s$ for the shared expert. The final prediction $\hat{\mathbf{y}}$ is the weighted sum of all expert outputs:
\begin{equation}
    \hat{\mathbf{y}} = w_s E_s(\mathbf{x}) + \sum_{m=1}^M w_m E_m(\mathbf{x})
\end{equation}
The router training objective minimizes the Mean Squared Error (MSE) loss between the prediction $\hat{\mathbf{y}}$ and the ground truth $\mathbf{y}$:
\begin{equation}
\mathcal{L}_{\text{MSE}} = \frac{1}{N} \left\| \hat{\mathbf{y}} - \mathbf{y} \right\|_2^2
\end{equation}

\begin{table*}[ht!]
\label{tab:main exp}
\centering
\small
\setlength{\heavyrulewidth}{1.2pt}  
\setlength{\lightrulewidth}{0.8pt}  
\setlength{\cmidrulewidth}{0.8pt}   
\setlength{\aboverulesep}{0.6ex}
\setlength{\belowrulesep}{0.8ex}
\setlength{\cmidrulesep}{0.8ex}
\setlength{\tabcolsep}{3.5pt}
\resizebox{0.95\textwidth}{!}{%
\begin{tabular}{@{}c|cccccccccccccccc@{}}
\toprule
{\textbf{Model}}                         & \multicolumn{2}{c}{\textbf{NEST(Ours)}}                                        & \multicolumn{2}{c}{\textbf{iTransformer}}                                    & \multicolumn{2}{c}{\textbf{UniTS}}                                     & \multicolumn{2}{c}{\textbf{PatchTST}}                                        & \multicolumn{2}{c}{\textbf{DLinear}}                             & \multicolumn{2}{c}{\textbf{FEDformer}}                           & \multicolumn{2}{c}{\textbf{Autoformer}}                          & \multicolumn{2}{c}{\textbf{Informer}}                        \\ \cmidrule(lr){2-3} \cmidrule(lr){4-5}  \cmidrule(lr){6-7}  \cmidrule(lr){8-9} \cmidrule(lr){10-11} \cmidrule(lr){12-13} \cmidrule(lr){14-15} \cmidrule(lr){16-17}
\multicolumn{1}{c|}{\textbf{Metric}}                       & \textbf{MSE}            & \multicolumn{1}{c|}{\textbf{MAE}}            & \textbf{MSE}         & \multicolumn{1}{c|}{\textbf{MAE}}         & \textbf{MSE}      & \multicolumn{1}{c|}{\textbf{MAE}}      & \textbf{MSE}         & \multicolumn{1}{c|}{\textbf{MAE}}         & \textbf{MSE}   & \multicolumn{1}{c|}{\textbf{MAE}}   & \textbf{MSE}   & \multicolumn{1}{c|}{\textbf{MAE}}   & \textbf{MSE}   & \multicolumn{1}{c|}{\textbf{MAE}}   & \textbf{MSE}   & \textbf{MAE}                    \\ \midrule
\multicolumn{1}{c|}{$\text{CESNET}_{1}$} & {\color{red}\textbf{0.294}} & \multicolumn{1}{c|}{{\color{red}\textbf{0.375}}} & 0.345       & \multicolumn{1}{c|}{0.398}       & 0.420    & \multicolumn{1}{c|}{0.427}    & {\color{blue}\underline{0.310}} & \multicolumn{1}{c|}{{\color{blue}\underline{0.383}}} & 1.143 & \multicolumn{1}{c|}{0.925} & 0.531 & \multicolumn{1}{c|}{0.553} & 0.699 & \multicolumn{1}{c|}{0.637} & 4.901 & 1.969                  \\ \midrule
\multicolumn{1}{c|}{$\text{CESNET}_{2}$} & {\color{red}\textbf{0.632}} & \multicolumn{1}{c|}{{\color{red}\textbf{0.559}}} & 0.676       & \multicolumn{1}{c|}{0.583}       & 0.729    & \multicolumn{1}{c|}{0.607}    & {\color{blue}\underline{0.654}} & \multicolumn{1}{c|}{{\color{blue}\underline{0.573}}} & 0.749 & \multicolumn{1}{c|}{0.688} & 0.842 & \multicolumn{1}{c|}{0.695} & 0.966 & \multicolumn{1}{c|}{0.765} & 3.119 & 1.496                  \\ \midrule
\multicolumn{1}{c|}{$\text{CESNET}_{3}$} & {\color{red}\textbf{0.334}} & \multicolumn{1}{c|}{{\color{red}\textbf{0.338}}} & 0.341       & \multicolumn{1}{c|}{0.342}       & 0.358    & \multicolumn{1}{c|}{0.351}    & {\color{blue}\underline{0.338}} & \multicolumn{1}{c|}{{\color{blue}\underline{0.340}}} & 0.343 & \multicolumn{1}{c|}{0.355} & 0.383 & \multicolumn{1}{c|}{0.380} & 0.389 & \multicolumn{1}{c|}{0.391} & 0.865 & 0.632                  \\ \midrule
\multicolumn{1}{c|}{$\text{CESNET}_{4}$} & {\color{red}\textbf{0.364}} & \multicolumn{1}{c|}{{\color{red}\textbf{0.405}}} & 0.406       & \multicolumn{1}{c|}{0.425}       & 0.446    & \multicolumn{1}{c|}{0.439}    & {\color{blue}\underline{0.382}} & \multicolumn{1}{c|}{{\color{blue}\underline{0.416}}} & 0.771 & \multicolumn{1}{c|}{0.725} & 0.452 & \multicolumn{1}{c|}{0.487} & 0.635 & \multicolumn{1}{c|}{0.601} & 2.222 & 1.312                  \\ \midrule
\multicolumn{1}{c|}{$\text{CESNET}_{5}$} & {\color{red}\textbf{0.682}} & \multicolumn{1}{c|}{{\color{red}\textbf{0.466}}} & 0.693       & \multicolumn{1}{c|}{{\color{blue}\underline{0.472}}} & 0.779    & \multicolumn{1}{c|}{0.511}    & {\color{blue}\underline{0.692}} & \multicolumn{1}{c|}{0.474}       & 0.691 & \multicolumn{1}{c|}{0.479} & 0.790 & \multicolumn{1}{c|}{0.531} & 0.806 & \multicolumn{1}{c|}{0.537} & 0.746 & 0.509                  \\ \midrule
\multicolumn{1}{c|}{Weather}   & {\color{red}\textbf{0.224}} & \multicolumn{1}{c|}{{\color{red}\textbf{0.262}}} & 0.231       & \multicolumn{1}{c|}{0.270}       & 0.308    & \multicolumn{1}{c|}{0.343}    & {\color{blue}\underline{0.229}} & \multicolumn{1}{c|}{{\color{blue}\underline{0.268}}} & 0.242 & \multicolumn{1}{c|}{0.294} & 0.329 & \multicolumn{1}{c|}{0.370} & 0.443 & \multicolumn{1}{c|}{0.470} & 0.704 & 0.597                  \\ \midrule
\multicolumn{1}{c|}{ETTh1}     & {\color{red}\textbf{0.515}} & \multicolumn{1}{c|}{{\color{red}\textbf{0.515}}} & {\color{blue}\underline{0.520}} & \multicolumn{1}{c|}{0.524}       & 0.994    & \multicolumn{1}{c|}{0.723}    & 0.523       & \multicolumn{1}{c|}{{\color{blue}\underline{0.518}}} & 0.526 & \multicolumn{1}{c|}{0.524} & 0.659 & \multicolumn{1}{c|}{0.621} & 0.817 & \multicolumn{1}{c|}{0.679} & 1.407 & 0.918                  \\ \midrule
\multicolumn{1}{c|}{ETTh2}     & {\color{blue}\underline{0.234}} & \multicolumn{1}{c|}{{\color{red}\textbf{0.334}}} & 0.236 & \multicolumn{1}{c|}{{\color{blue}\underline{0.338}}} & 0.297    & \multicolumn{1}{c|}{0.386}    & 0.251       & \multicolumn{1}{c|}{0.347}       & {\color{red}\textbf{0.229}} & \multicolumn{1}{c|}{{\color{blue}\underline{0.338}}} & 0.315 & \multicolumn{1}{c|}{0.424} & 0.465 & \multicolumn{1}{c|}{0.511} & 0.536 & 0.539                  \\ \midrule
\multicolumn{1}{c|}{TEC}      & {\color{red}\textbf{0.691}} & \multicolumn{1}{c|}{{\color{red}\textbf{0.574}}} & {\color{blue}\underline{0.694}} & \multicolumn{1}{c|}{{\color{blue}\underline{0.577}}} & 0.803 & \multicolumn{1}{c|}{0.642} & {\color{blue}\underline{0.694}} & \multicolumn{1}{c|}{{\color{blue}\underline{0.577}}} & 0.695 & \multicolumn{1}{c|}{0.581} & 0.735 & \multicolumn{1}{c|}{0.603} & 0.869 & \multicolumn{1}{c|}{0.667} & 0.723 & 0.594                  \\ \midrule
\multicolumn{1}{c|}{\textbf{$1^{st}$ Count}} & {\color{red}\textbf{32}}             & \multicolumn{1}{c|}{{\color{red}\textbf{32}}}             & 1           & \multicolumn{1}{c|}{3}           & 0        & \multicolumn{1}{c|}{0}        & 3           & \multicolumn{1}{c|}{{\color{blue}\underline{4}}}          & {\color{blue}\underline{4}}     & \multicolumn{1}{c|}{0}     & 0     & \multicolumn{1}{c|}{0}     & 0     & \multicolumn{1}{c|}{0}     & 0     & {0} \\ \bottomrule
\end{tabular} 
}
\caption{Long term forecasting results of time series models on nine different datasets. The history sequence length is 512. Corresponding prediction lengths include \{96, 192, 336, 720\}. A lower MSE or MAE indicates a better prediction. Averaged results of four prediction lengths are reported here. $1^{st}$ Count represents the number of wins achieved by a model under all prediction lengths and datasets. The best result is {\color{red}\textbf{bolded}} and the second best is {\color{blue}\underline{underlined}}.}
\end{table*}

\begin{table*}[ht!]

\label{tab:ablation}
\centering
\small
\setlength{\heavyrulewidth}{1.2pt}
\setlength{\lightrulewidth}{0.8pt}
\setlength{\cmidrulewidth}{0.8pt}
\setlength{\aboverulesep}{0.6ex}
\setlength{\belowrulesep}{0.8ex}
\setlength{\tabcolsep}{2.5pt}

\resizebox{\textwidth}{!}{%
\begin{tabular}{l|cc|cc|cc|cc|cc|cc|cc|cc|cc}
\toprule
\textbf{Dataset} & \multicolumn{2}{c}{\textbf{$\text{CESNET}_{1}$}} & \multicolumn{2}{c}{\textbf{$\text{CESNET}_{2}$}} & \multicolumn{2}{c}{\textbf{$\text{CESNET}_{3}$}} & \multicolumn{2}{c}{\textbf{$\text{CESNET}_{4}$}} & \multicolumn{2}{c}{\textbf{$\text{CESNET}_{5}$}} & \multicolumn{2}{c}{\textbf{Weather}} & \multicolumn{2}{c}{\textbf{ETTh1}} & \multicolumn{2}{c}{\textbf{ETTh2}} & \multicolumn{2}{c}{\textbf{TEC}} \\ 
\cmidrule(lr){2-3} \cmidrule(lr){4-5} \cmidrule(lr){6-7} \cmidrule(lr){8-9} \cmidrule(lr){10-11} \cmidrule(lr){12-13} \cmidrule(lr){14-15} \cmidrule(lr){16-17} \cmidrule(lr){18-19}
\textbf{Metric} & \textbf{MSE} & \textbf{MAE} & \textbf{MSE} & \textbf{MAE} & \textbf{MSE} & \textbf{MAE} & \textbf{MSE} & \textbf{MAE} & \textbf{MSE} & \textbf{MAE} & \textbf{MSE} & \textbf{MAE} & \textbf{MSE} & \textbf{MAE} & \textbf{MSE} & \textbf{MAE} & \textbf{MSE} & \textbf{MAE} \\ \midrule

NEST & {\color{red}\textbf{0.294}} & {\color{red}\textbf{0.375}} & {\color{red}\textbf{0.632}} & {\color{red}\textbf{0.559}} & {\color{red}\textbf{0.334}} & {\color{red}\textbf{0.338}} & {\underline{0.364}} & {\underline{0.405}} & {\color{red}\textbf{0.682}} & {\color{red}\textbf{0.466}} & {\color{red}\textbf{0.224}} & {\color{red}\textbf{0.262}} & {\color{red}\textbf{0.515}} & {\color{red}\textbf{0.515}} & {\color{red}\textbf{0.234}} & {\underline{0.334}} & {\color{red}\textbf{0.691}} & {\color{red}\textbf{0.574}} \\

\textit{w/o Router} & 0.315 & {\underline{0.376}} & 0.656 & 0.572 & 0.336 & 0.340 & 0.400 & 0.417 & 0.727 & 0.492 & 0.244 & 0.281 & 0.699 & 0.609 & 0.242 & 0.343 & 0.698 & 0.585 \\

\textit{w/o Kmeans} & {\underline{0.307}} & 0.383 & 0.652 & 0.570 & 0.338 & 0.340 & {\color{red}\textbf{0.363}} & 0.406 & {\underline{0.687}} & {\underline{0.468}} & {\underline{0.225}} & {\underline{0.265}} & {\underline{0.519}} & {\underline{0.519}} & 0.246 & 0.342 & {\underline{0.692}} & {\underline{0.575}} \\

\textit{w/o R\&K} & 0.330 & 0.386 & 0.664 & 0.578 & 0.338 & 0.341 & 0.380 & {\color{red}\textbf{0.402}} & 0.766 & 0.506 & 0.268 & 0.307 & 0.630 & 0.579 & 0.252 & 0.353 & 0.700 & 0.586 \\

\textit{Dist. Router} & 0.309 & 0.380 & {\underline{0.637}} & {\underline{0.561}} & {\underline{0.335}} & {\underline{0.338}} & 0.392 & 0.415 & 0.709 & 0.480 & 0.230 & 0.265 & 0.522 & 0.521 & {\underline{0.234}} & {\color{red}\textbf{0.334}} & 0.699 & 0.581 \\

\bottomrule
\end{tabular}%
}
\caption{Ablation study on NEST components across main experiment datasets. The history sequence length is 512. Corresponding prediction lengths include \{96, 192, 336, 720\}. A lower MSE or MAE indicates a better prediction. Averaged results of four prediction lengths are reported here. The best result is {\color{red}\textbf{bolded}} and the second best is \underline{underlined}.}
\end{table*}

\section{Experiments}
\label{sec:experiments}

\subsection{Experimental Setup}
\label{ssec:exp_setup}


\textbf{Datasets.} To evaluate NEST's ability to capture evolving variable dependencies across diverse domains, we conduct experiments on several large-scale benchmarks: CESNET-TIMESERIES24~\cite{koumar2025cesnet}: A high-fidelity network traffic dataset containing 40 weeks of activity from over 275,000 IPs on an ISP network. Its high variability and heterogeneous behavioral modes present an authentic challenge for regime identification. We select five representative IP series for primary evaluation. Ionospheric TEC: A specialized physical dataset characterizing Earth's ionospheric dynamics which features decadal solar cycles and complex spatiotemporal relations. Public Benchmarks: To demonstrate generalizability, we include the Weather and ETT datasets(ETTh1, ETTh2), which are standard benchmarks for evaluating long-term forecasting robustness under varying non-stationarity. 

\textbf{Baselines.} We compare our proposed NEST model against a comprehensive suite of SOTA and classical deep learning models for time series forecasting. The baselines include the unified foundation model UniTS\cite{gao2024units}; Transformer-based models such as iTransformer\cite{liu2023itransformer}, PatchTST\cite{nie2022time}, FEDformer\cite{zhou2022fedformer}, Autoformer\cite{wu2021autoformer}, and Informer\cite{zhou2021informer}; and the MLP-based DLinear\cite{zeng2023transformers}. This diverse set of models allows for a rigorous comparison against different architectural paradigms. We perform hyperparameter search for a fair comparison. 

\textbf{Implementation Details.} To ensure a fair comparison, all models were trained and tested on a single NVIDIA V100 32GB GPU. We adopt the same strict chronological split for all benchmarking datasets: 70\% for training, 10\% for validation, and 20\% for testing. We conduct our main experiments with a fixed lookback window of 512 time steps.

\subsection{Main Results and Analysis}
\label{ssec:main_results}

The main forecasting results are presented in Table~\ref{tab:main exp}, which provides a comprehensive comparison of NEST against all baseline models across nine datasets and four prediction horizons. The analysis of these results highlights the significant and quantifiable advantages of our proposed architecture in handling complex, non-stationary long-span time series datasets.

\textbf{Superiority Across Diverse Benchmarks.} NEST consistently achieves state-of-the-art (SOTA) performance, substantially outperforming all competitive baselines. As indicated in the ``$1^{st}$ Count'' row of Table~\ref{tab:main exp}, NEST secures the best MSE and MAE results in 32 out of 36 experimental settings across 9 distinct datasets. The performance gains are particularly pronounced on high-variability datasets; for instance, on the $\text{CESNET}_1$ benchmark, NEST surpasses the strong baseline PatchTST by 5.49\% and iTransformer by 14.82\% in MSE on average across four different prediction horizons. This consistent dominance across multi-year physical cycles and heterogeneous network traffic validates NEST's capability to effectively track and model the dynamic reorganization of inter-variable dependencies, which is a structural challenge that monolithic models and local non-stationarity mitigators fail to address.

\textbf{Robustness on Heterogeneous Network Traffic.} NEST's efficacy is most prominent on the five large-scale \textbf{CESNET} benchmarks. Characterized by 40 weeks of real-world ISP traffic, these datasets exhibit extreme variability and regime transitions. NEST’s superiority here stems from its systematic handling of non-stationarity: the unsupervised regime discovery module utilizes moment-entropy metrics to categorize data slices into functionally distinct modes, such as high-volatility surges versus stable, low-complexity periods. 


\subsection{Ablation Study on NEST Components}
\label{ssec:ablation_core}
To quantify the individual contributions of the core components in NEST, we evaluate five variants: (1) the full \textbf{NEST} model; (2) \textit{w/o Router}, which replaces the regime-oriented router with a simple parameter averaging of experts; (3) \textit{w/o Kmeans}, which replaces the moment-entropy based clustering with naive sequential data partitioning; (4) \textit{w/o R\&K}, which removes both modules; and (5) \textit{Dist. Router}, which utilizes only the distance-based modulation $\tilde{\omega}_m$ without the temporal initial weight score $\mathbf{w}_{\text{init}}$. The results in Table~\ref{tab:ablation} demonstrate that the full NEST configuration achieves superior performance in nearly all metrics, validating the design of our regime-aware paradigm.

\textbf{Efficacy of the Regime-Oriented Router:} Removing the router(\textit{w/o Router}) leads to a consistent performance drop, with MSE increasing markedly on datasets like ETTh1 (0.515 to 0.699). This underlines that simple expert averaging fails to capture the structural transitions between regimes. Furthermore, the performance gap between NEST and \textit{Dist. Router} confirms the necessity of our routing strategy: while geometric distance provides a global anchor, the initial weight score $w_{\text{init}}$ is crucial for identifying the temporal state of the system.

\textbf{Necessity of Moment-Entropy Based Discovery:} The \textit{w/o Kmeans} variant, which lacks principled regime partitioning, underperforms the full model across diverse benchmarks. For instance, on ETTh2, the MSE increases from 0.234 to 0.246. This justifies our use of moment-entropy metrics as a sufficient mathematical basis to discover functionally distinct operational modes, as naive partitioning fails to ensure the structural specialization of experts.

\textbf{Synergistic Effect and Robustness:} The \textit{w/o R\&K} variant generally yields the poorest results, highlighting the synergy between regime discovery and adaptive routing. The overwhelming trend across datasets confirms that the integration of principled regime discovery and regime-oriented routing is vital for robust forecasting under structural distribution shifts.

\begin{figure}[ht!]
    \centering  
    \includegraphics[width=0.5\textwidth]{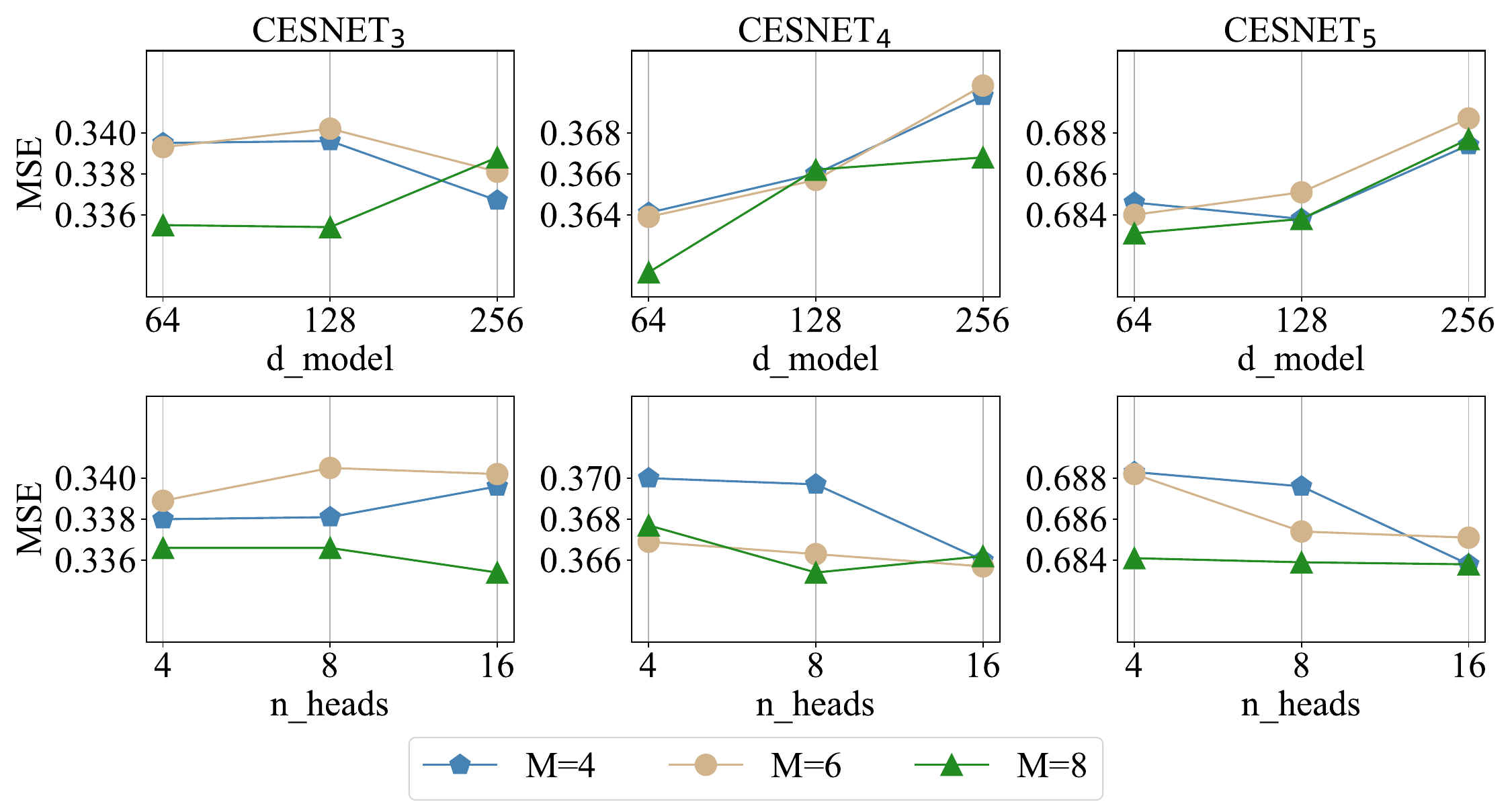}
    \caption{\textbf{Hyperparameter sensitivity analysis of NEST.} 
     Parameter sensitivity analysis of NEST across multiple CESNET subsets. The plots illustrate the impact of hidden dimension size ($d_{model}$), the number of attention heads ($n_{heads}$), and the number of discovered regimes ($M$) on forecasting performance (MSE).}
    \label{fig:hyperparam_sensitivity}
\end{figure}

\subsection{Hyperparameter Sensitivity Analysis}
\label{ssec:hyperparam_sensitivity}

We evaluate NEST's robustness against key hyperparameters: model dimension ($d_{model}$), attention heads ($n_{heads}$), and regime count ($M$). As shown in Figure~\ref{fig:hyperparam_sensitivity}, NEST maintains remarkable stability across varying architectural configurations. Specifically, performance remains consistent for $d_{model} \in \{64, 128\}$ and $n_{heads} \in \{4, 8, 16\}$, with slight MSE increases at $d_{model}=256$. This insensitivity suggests that NEST's efficacy is derived from its regime-aware logic rather than exhaustive parameter tuning. In contrast, the regime count $M$ (Data Slices) exhibits a more data-dependent impact. Empirical results across multiple benchmarks suggest that $M=8$ serves as a robust and versatile configuration, providing a sufficient hypothesis space for experts to capture the diverse behavioral modes within the moment-entropy space. At the same time, we find that a smaller $M$ (e.g., $M=4$) can occasionally yield competitive results on datasets with relatively low regime complexity. This indicates that while NEST benefits from a diverse expert pool to handle high dataset-level distribution shifts, it maintains a degree of parameter efficiency, allowing for a pruned architecture when the underlying system transitions are less intricate.




\begin{table}[t]
\centering
\label{tab:cka permutation}
\renewcommand{\arraystretch}{1.1} 
\small
\resizebox{0.45\textwidth}{!}{%
\begin{tabular}{cccccc|cc}
\toprule
\textbf{Model} & \textbf{Metric} & \textbf{96} & \textbf{192} & \textbf{336} & \textbf{720} & \textbf{Avg.} \\ \midrule
 & MSE & \textbf{0.149} & \textbf{0.192} & \textbf{0.243} & \textbf{0.311} & \textbf{0.224} \\
\multirow{-2}{*}{\textbf{NEST}} & MAE & \textbf{0.199} & \textbf{0.242} & \textbf{0.279} & \textbf{0.327} & \textbf{0.262} \\ \midrule
 & MSE & 0.156 & 0.223 & 0.250 & 0.322 & 0.237 \\
\multirow{-2}{*}{\textbf{CKA-Swap}} & MAE & 0.206 & 0.265 & 0.282 & 0.336 & 0.272 \\ 
\bottomrule
\end{tabular}
}
\caption{Expert weight permutation analysis based on CKA similarity. Prediction lengths include \{96, 192, 336, 720\}.}
\end{table}

\subsection{Study on Regime Experts}
\label{ssec:interpretability}

To investigate the internal mechanism of NEST, we conduct a multi-level interpretability analysis to verify the structural specialization of the expert pool and the efficacy of the regime-oriented router.

\textbf{Visualizing Variate-Attention Map.} We first visualize the Query-Key (QK) attention matrices of different regime experts to observe their focus on inter-variable dependencies. The attention patterns across experts exhibit heterogeneity, especially in the high attention score QK-pair vicinity. Each expert evolves a distinct variate-attention map that adapts to the input sample, representing a specific coupling logic of variables. This polymorphism confirms that experts have effectively specialized in different system dynamics rather than converging to a redundant averaged representation.

\begin{figure}[t]
    \centering  
    \includegraphics[width=0.38\textwidth]{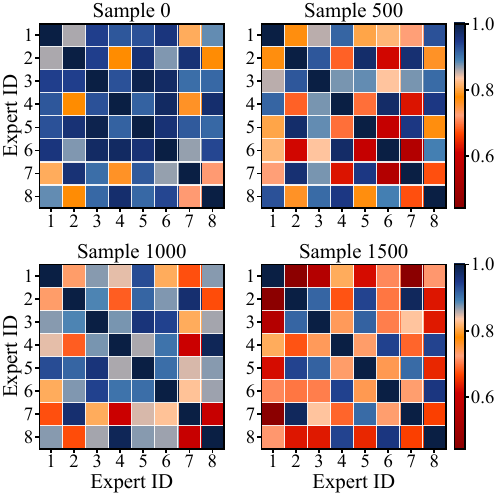}
    \caption{CKA similarity heatmap of specialized experts' QK matrix. Low pairwise similarity scores indicate lower similarity between the experts. Results are sampled from the Weather test set.}
    \label{fig:cka}
\end{figure}

\textbf{Quantifying Expert Dissimilarity via CKA.} To further quantify the functional divergence between experts, we employ Centered Kernel Alignment (CKA) to measure the similarity of their attention matrices. As shown in Figure~\ref{fig:cka}, the low pairwise CKA scores across the expert pool indicate significant representational dissimilarity. This high degree of heterogeneity proves that our moment-entropy based regime discovery successfully facilitates the emergence of non-redundant, specialized experts.

\textbf{Expert Permutation and Router Sensitivity.} To validate the necessity of precise expert orchestration, we conduct a weight permutation experiment. For each batch sample, experts with CKA similarity above a predefined threshold are eligible for pairing and routing score swapping, where pairs with higher similarity are prioritized for the swap operation. The results, summarized in Table~\ref{tab:cka permutation}, show a significant performance degradation across all test samples. Even when two experts appear relatively similar, the router's precise allocation remains critical. This ``collapse" upon permutation demonstrates that the efficacy of NEST stems from the exact match between the detected regime and the expert’s specialized variate-attention kernel, rather than mere parameter ensemble.

\section{Conclusion}
In this paper, we addressed the challenge of dataset-level distribution shifts by introducing NEST, a two-phase MoE framework that models the alternation of underlying regimes. By combining moment-entropy based regime discovery, structurally specialized experts, and a regime-oriented router, NEST decomposes heterogeneous temporal dynamics into reusable forecasting kernels and adaptively recomposes them for each input. Extensive evaluations demonstrate that NEST achieves SOTA performance across diverse benchmarks, particularly on highly non-stationary long-term datasets. 

\clearpage

\section*{Acknowledgments}
This work was supported by the grants from the Natural Science Foundation of China (U2541212, 62572035), and Beijing Natural Science Foundation (L248032). Thanks for the computing infrastructure provided by Beijing Advanced Innovation Center for Big Data and Brain Computing. Haoyi Zhou is the corresponding author.

\bibliographystyle{named}
\bibliography{ijcai26}

@inproceedings{zhou2021informer,
  title={Informer: Beyond efficient transformer for long sequence time-series forecasting},
  author={Zhou, Haoyi and Zhang, Shanghang and Peng, Jieqi and Zhang, Shuai and Li, Jianxin and Xiong, Hui and Zhang, Wancai},
  booktitle={Proceedings of the AAAI conference on artificial intelligence},
  volume={35},
  number={12},
  pages={11106--11115},
  year={2021}
}

@inproceedings{liu2021pyraformer,
  title={Pyraformer: Low-complexity pyramidal attention for long-range time series modeling and forecasting},
  author={Liu, Shizhan and Yu, Hang and Liao, Cong and Li, Jianguo and Lin, Weiyao and Liu, Alex X and Dustdar, Schahram},
  booktitle={International conference on learning representations},
  year={2021}
}

@inproceedings{zeng2023transformers,
  title={Are transformers effective for time series forecasting?},
  author={Zeng, Ailing and Chen, Muxi and Zhang, Lei and Xu, Qiang},
  booktitle={Proceedings of the AAAI conference on artificial intelligence},
  volume={37},
  number={9},
  pages={11121--11128},
  year={2023}
}

@article{wu2021autoformer,
  title={Autoformer: Decomposition transformers with auto-correlation for long-term series forecasting},
  author={Wu, Haixu and Xu, Jiehui and Wang, Jianmin and Long, Mingsheng},
  journal={Advances in Neural Information Processing Systems},
  volume={34},
  pages={22419--22430},
  year={2021}
}

@article{boehme2007applying,
  title={Applying time series to power flow analysis in networks with high wind penetration},
  author={Boehme, Thomas and Wallace, A Robin and Harrison, Gareth P},
  journal={IEEE transactions on power systems},
  volume={22},
  number={3},
  pages={951--957},
  year={2007},
  publisher={IEEE}
}

@article{pulkkinen2007space,
  title={Space weather: Terrestrial perspective},
  author={Pulkkinen, Tuija},
  journal={Living Reviews in Solar Physics},
  volume={4},
  number={1},
  pages={1},
  year={2007},
  publisher={Springer}
}

@inproceedings{basu1996time,
  title={Time series models for internet traffic},
  author={Basu, Sabyasachi and Mukherjee, Amarnath and Klivansky, Steve},
  booktitle={Proceedings of IEEE INFOCOM'96. Conference on Computer Communications},
  volume={2},
  pages={611--620},
  year={1996},
  organization={IEEE}
}

@inproceedings{zhou2022fedformer,
  title={Fedformer: Frequency enhanced decomposed transformer for long-term series forecasting},
  author={Zhou, Tian and Ma, Ziqing and Wen, Qingsong and Wang, Xue and Sun, Liang and Jin, Rong},
  booktitle={International Conference on Machine Learning},
  pages={27268--27286},
  year={2022},
  organization={PMLR}
}

@article{wu2020deep,
  title={Deep transformer models for time series forecasting: The influenza prevalence case},
  author={Wu, Neo and Green, Bradley and Ben, Xue and O'Banion, Shawn},
  journal={arXiv preprint arXiv:2001.08317},
  year={2020}
}

@article{nie2022time,
  title={A time series is worth 64 words: Long-term forecasting with transformers},
  author={Nie, Yuqi and Nguyen, Nam H and Sinthong, Phanwadee and Kalagnanam, Jayant},
  journal={arXiv preprint arXiv:2211.14730},
  year={2022}
}

@article{cleveland1990stl,
  title={STL: A seasonal-trend decomposition},
  author={Cleveland, Robert B and Cleveland, William S and McRae, Jean E and Terpenning, Irma},
  journal={J. Off. Stat},
  volume={6},
  number={1},
  pages={3--73},
  year={1990}
}

@article{torrence1998practical,
  title={A practical guide to wavelet analysis},
  author={Torrence, Christopher and Compo, Gilbert P},
  journal={Bulletin of the American Meteorological society},
  volume={79},
  number={1},
  pages={61--78},
  year={1998},
  publisher={American Meteorological Society}
}

@article{liu2023itransformer,
  title={itransformer: Inverted transformers are effective for time series forecasting},
  author={Liu, Yong and Hu, Tengge and Zhang, Haoran and Wu, Haixu and Wang, Shiyu and Ma, Lintao and Long, Mingsheng},
  journal={arXiv preprint arXiv:2310.06625},
  year={2023}
}

@inproceedings{siami2018comparison,
  title={A comparison of ARIMA and LSTM in forecasting time series},
  author={Siami-Namini, Sima and Tavakoli, Neda and Namin, Akbar Siami},
  booktitle={2018 17th IEEE international conference on machine learning and applications (ICMLA)},
  pages={1394--1401},
  year={2018},
  organization={Ieee}
}

@article{weerakody2021review,
  title={A review of irregular time series data handling with gated recurrent neural networks},
  author={Weerakody, Philip B and Wong, Kok Wai and Wang, Guanjin and Ela, Wendell},
  journal={Neurocomputing},
  volume={441},
  pages={161--178},
  year={2021},
  publisher={Elsevier}
}

@inproceedings{kim2021reversible,
  title={Reversible instance normalization for accurate time-series forecasting against distribution shift},
  author={Kim, Taesung and Kim, Jinhee and Tae, Yunwon and Park, Cheonbok and Choi, Jang-Ho and Choo, Jaegul},
  booktitle={International conference on learning representations},
  year={2021}
}

@article{ye2024frequency,
  title={Frequency adaptive normalization for non-stationary time series forecasting},
  author={Ye, Weiwei and Deng, Songgaojun and Zou, Qiaosha and Gui, Ning},
  journal={Advances in Neural Information Processing Systems},
  volume={37},
  pages={31350--31379},
  year={2024}
}

@inproceedings{ogasawara2010adaptive,
  title={Adaptive normalization: A novel data normalization approach for non-stationary time series},
  author={Ogasawara, Eduardo and Martinez, Leonardo C and De Oliveira, Daniel and Zimbr{\~a}o, Geraldo and Pappa, Gisele L and Mattoso, Marta},
  booktitle={The 2010 International Joint Conference on Neural Networks (IJCNN)},
  pages={1--8},
  year={2010},
  organization={IEEE}
}

@article{passalis2019deep,
  title={Deep adaptive input normalization for time series forecasting},
  author={Passalis, Nikolaos and Tefas, Anastasios and Kanniainen, Juho and Gabbouj, Moncef and Iosifidis, Alexandros},
  journal={IEEE transactions on neural networks and learning systems},
  volume={31},
  number={9},
  pages={3760--3765},
  year={2019},
  publisher={IEEE}
}

@article{liu2023adaptive,
  title={Adaptive normalization for non-stationary time series forecasting: A temporal slice perspective},
  author={Liu, Zhiding and Cheng, Mingyue and Li, Zhi and Huang, Zhenya and Liu, Qi and Xie, Yanhu and Chen, Enhong},
  journal={Advances in Neural Information Processing Systems},
  volume={36},
  pages={14273--14292},
  year={2023}
}

@inproceedings{fan2023dish,
  title={Dish-ts: a general paradigm for alleviating distribution shift in time series forecasting},
  author={Fan, Wei and Wang, Pengyang and Wang, Dongkun and Wang, Dongjie and Zhou, Yuanchun and Fu, Yanjie},
  booktitle={Proceedings of the AAAI conference on artificial intelligence},
  volume={37},
  number={6},
  pages={7522--7529},
  year={2023}
}

@article{gao2024units,
  title={Units: A unified multi-task time series model},
  author={Gao, Shanghua and Koker, Teddy and Queen, Owen and Hartvigsen, Tom and Tsiligkaridis, Theodoros and Zitnik, Marinka},
  journal={Advances in Neural Information Processing Systems},
  volume={37},
  pages={140589--140631},
  year={2024}
}

@article{koumar2025cesnet,
  title={CESNET-TimeSeries24: Time Series Dataset for Network Traffic Anomaly Detection and Forecasting},
  author={Koumar, Josef and Hynek, Karel and {\v{C}}ejka, Tom{\'a}{\v{s}} and {\v{S}}i{\v{s}}ka, Pavel},
  journal={Scientific Data},
  volume={12},
  number={1},
  pages={338},
  year={2025},
  publisher={Nature Publishing Group UK London}
}

\end{document}